# Can ML predict the solution value for a difficult combinatorial problem?

Constantine Goulimis, Gastón Simone

Greycon Ltd., 7 Calico House, Plantation Wharf, London SW11 3TN.
{cng,gs}@greycon.com

**Abstract**

We look at whether machine learning ('ML') can predict the final objective function value of a difficult combinatorial optimisation problem from the input. Our context is the pattern reduction problem, one industrially important but difficult aspect of the cutting stock problem. Machine learning appears to have higher prediction accuracy than a naïve model, reducing mean absolute percentage error ('MAPE') from 12.0% to 8.7%.

## 1 Background & related work

The many practical applications of the one-dimensional cutting stock problem ('1D-CSP') have provided a rich source of challenges to the mathematical optimisation community. In this paper we look at one such aspect, namely the minimisation of cutting patterns within the universe of minimum waste solutions.

It is well known that the 1D-CSP is quite degenerate, i.e. multiple different solutions with the same waste often exist. This can be explained by geometrical re-arrangement, i.e. it is sometimes possible, for example, to swap items belonging to different patterns, creating new patterns in the process.

In some industrial settings, particularly in the plastic film industry, pattern minimisation is quite important because the technical characteristics of the slitter winders that cut the material are such that changing patterns can cause production bottlenecks.

In general, depending on the machinery, the actual effort in producing a particular 1D-CSP solution is a multi-faceted problem; pattern minimisation is an important aspect, but not the only one. Sometimes the time cost of a pattern change is not fixed, but depends on the differences from the previous one. For example, the sequencing of a given set of patterns and the relative position of the rolls in each pattern gives rise to the knife change minimisation problem, which can be modelled as a generalised travelling salesman problem. Nonetheless, pattern count has become a common key performance indicator.

See [1] for an overview of different approaches for addressing the pattern minimisation problem. It is known that pattern minimisation is NP-hard [2]. In practice, this might not mean much, [3]. However, it is harder than the waste-minimisation 1D-CSP, for which optimal integer solutions have been known since 1990, [4]. It is definitely not a well-solved problem, in the sense that in the real world we encounter instances that are not solvable to optimality with the typical computational budget that is available (a few minutes). For example, [5] and [6], (but note that [7] cast doubts on the validity of the results of the latter) employed time limits of 2 and ½ hours respectively. In addition, both of these methods fail to include real-world constraints, such as position constraints or number of knives (and there are many others).

In our X-Trim commercial application, we use a combination of *transformation heuristics* [1] complemented by semi-exact methods based on integer programming using Gurobi 8.1 as the solver. Allowing this combined algorithm to run to completion for mid- and large-sized instances may take a very long time (hours) and therefore users tend to interrupt the search after a few minutes.

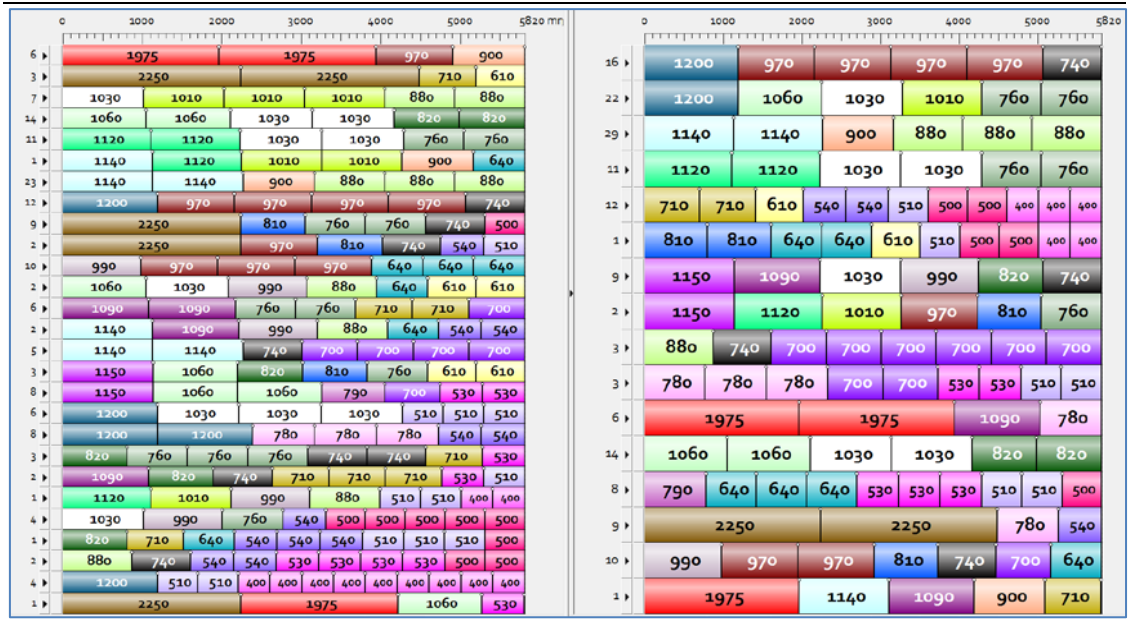

**Figure 1. Solutions to an instance with 29 distinct sizes. The solution on the left has 27 patterns and an equivalent solution with 16 patterns is on the right. It is not known whether equivalent solutions with fewer patterns exist.**

The problem we address in this paper is whether we can predict, from the starting solution, the final outcome, in terms of the number of patterns. In this quest we have three aims: Firstly, we would like to provide quick guidance to the planner as to the likely outcome; without such guidance she is restricted to watching the optimisation and, at some point, either accepting any solution found so far or cancelling. Secondly, we can also use the ML estimate as a stopping criterion. Thirdly, we are aiming to learn a meaningful property of this particular optimisation problem (see section 3.2.2 of [8]) and, in particular, whether machine learning (ML) algorithms can discern structure, which is otherwise invisible even to the expert. A positive outcome to the latter aim raises some obvious interesting possibilities.

We define some terminology: suppose we have a solution $\omega$ to an instance of a 1D-CSP, where the items are of size $l_i \in (0,1)$ (expressed as a fraction of the master size). The solution is completely represented as a set of pairs. Each pair consists of two elements, the number of repetitions of the pattern and its contents – the latter is a multi-set $\langle l_i \rangle$ of the required sizes. For example, the first element of the solution shown to the right of Figure 1 is:

$$(16, \langle \frac{1200}{5820}, \frac{970}{5820}, \frac{970}{5820}, \frac{970}{5820}, \frac{970}{5820}, \frac{740}{5820} \rangle)$$

To avoid trivial cases where the same pattern repeats, we require that the multi-sets are unique. The cardinality of the solution (number of elements in the set of pairs) is the pattern count. We call two solutions *equivalent*, if (a) the solution *run length* (sum of the pattern repetitions) is the same and (b) the total production for each item $l_i$ is the same. In Figure 1, the solution run length is 156 and, e.g., both solutions contain 30 pieces of size 900.

Define $f: \Omega \mapsto \mathbb{N}$, to be the function that takes a solution $\omega \in \Omega$ and returns the minimum number of patterns for that solution. This is convex in the sense that if we split a solution into two parts $\omega_1, \omega_2: \omega_1 \cap \omega_2 = \emptyset \,\&\, \omega_1 \cup \omega_2 = \omega$, then $f(\omega) \leq f(\omega_1) + f(\omega_2)$. However, we cannot effectively calculate $f$, only an upper bound $\bar{f}$; this is not necessarily convex.

## 2  Data Generation & Scaling

The training set in a ML context typically should consist of several thousand instances. We therefore constructed a random instance generator. Since the problems encountered in industry are far from uniform, we created instances for three families:

| Family | Item Width | Primary Width | Comment |
| --- | --- | --- | --- |
| **Corrugated Case Materials (CCM)** | typical widths of the reels are in the range 1800-2500 mm, typically in multiples of 10 or 25 mm, with a greater frequency for the larger widths. Widths both below and above this range also occur, but less frequently. | The paper machines tend to be 5-8 m wide | CCM accounts for ~⅓ of global total paper production – output goes into producing the familiar brown cardboard boxes. 1,500 instances. |
| **Plastic Film (F)** | majority of sizes are in the 300-1000 mm size (in multiples of 5 mm), but there are also parent reels in the range 1800-2300 mm for metallised film | primary process (extruder) is typically 6-8 m wide | These represent the production of polypropylene film, widely used in food packaging. 6,800 instances. |
| **Fine paper (FP)** | parent reels for sheeting, in the range 1500-2300 mm | paper machines in the range 4-6 m | These represent the production of paper sheets. 1,000 instances. |

Because of the importance of pattern count in the film industry, we deliberately over-represented this in the instances we generated.

Using the data generator we created 9,300 instances. For each of those we had to also create the initial minimum waste solution and then, using the pattern reduction algorithm with a time budget of 150 seconds, the reduced solution. The process took ~800 hours on a dedicated 6-core computer (processor: Intel® Core™ i7-6800K @ 4.30 GHz, RAM: 16 GB, OS: 64-bit Windows 10 Enterprise). The dataset is available upon request from the authors.

One of the complications with this data is that we do not actually know $f(\omega)$, the minimum number of patterns for an instance $\omega$; instead we have an upper bound $\bar{f}(\omega)$, obtained from our imperfect algorithm. This becomes relevant when discussing the accuracy of a predictor. We noticed that some of the worst absolute percentage errors (difference between forecast and actual) occur in the ~3% of instances where the pattern reduction algorithm achieves no improvement within the time budget. Yet, with some manual intervention (using the convexity property), it is possible to coax reductions in all such instances. However, we did not use these manually-improved solutions in the comparison, because we felt it would be more useful to predict the benefit the user might obtain (with a time budget of 150 seconds) rather than what is achievable in an ideal scenario. This makes this analysis sensitive to changes to the pattern reduction algorithm; as and when algorithm improvements are made, the training will have to be re-done.

In presenting the solution to the ML algorithm, one issue is how to represent solutions with a different initial number of patterns (they range from 5 to 66 in our training data set). This is analogous to image recognition with images of different size (see [9]).

In our case, we decided on a simple form:

$$\begin{pmatrix} c_1 & n_{1,1} & w_{1,1} & \cdots & n_{1,k} & w_{1,k} \\ \vdots & \vdots & \vdots & \ddots & \vdots & \vdots \\ c_p & n_{p,1} & w_{p,1} & \cdots & n_{p,k} & w_{p,k} \\ \vdots & \vdots & \vdots & \ddots & \vdots & \vdots \\ c_M & n_{M,1} & w_{M,1} & \cdots & n_{M,k} & w_{M,k} \end{pmatrix}$$

This matrix, of size $M \times (1 + 2k)$, represents each pattern in the solution as one of its rows, containing the following information:

- $c_p$: number of repetitions of pattern $p$
- $n_{p,i}$: number of repetitions of item $i$ in pattern $p$
- $w_{p,i}$: width of item $i$ in pattern $p$

Parent reels in a pattern and patterns in the solution are ordered, decreasingly, by widths. We use *M=400* and *k=12*, as these dimensions are enough for all problems we have found in real scenarios. Smaller solutions are padded with zeros to get a matrix shaped as above.

So, the first pattern in the solution to the right of Figure 1, corresponding to the first row of the above matrix, is represented as:

$$\begin{pmatrix} 16 & 1 & \frac{1200}{5820} & 4 & \frac{970}{5820} & 1 & \frac{740}{5820} & 0 & \cdots & 0 \end{pmatrix}$$

We experimented with shrinking the values of *M* and *k* to smaller values, since the vast majority of real-world instances (including all in the dataset we have generated) have $M \leq 80$ and $k \leq 6$. Although there was a small difference in the (anyway small) training time, there was no statistically significant difference in the results. We suspect this is because ML technology has become quite good at identifying 'empty' information.

The solution representation removes some of the natural redundancy (e.g. changing the sequence of items within a pattern). In the process, we found that different solution representations have a big impact on the learning speed.

## 3 The Naïve Model

The naïve model compares the initial and final pattern counts and fits a quadratic on 80% (7,440) randomly-chosen instances of the 9,300-instance collection; we find it quite astonishing that this naïve model has an $R^2$ value of 86.8%:

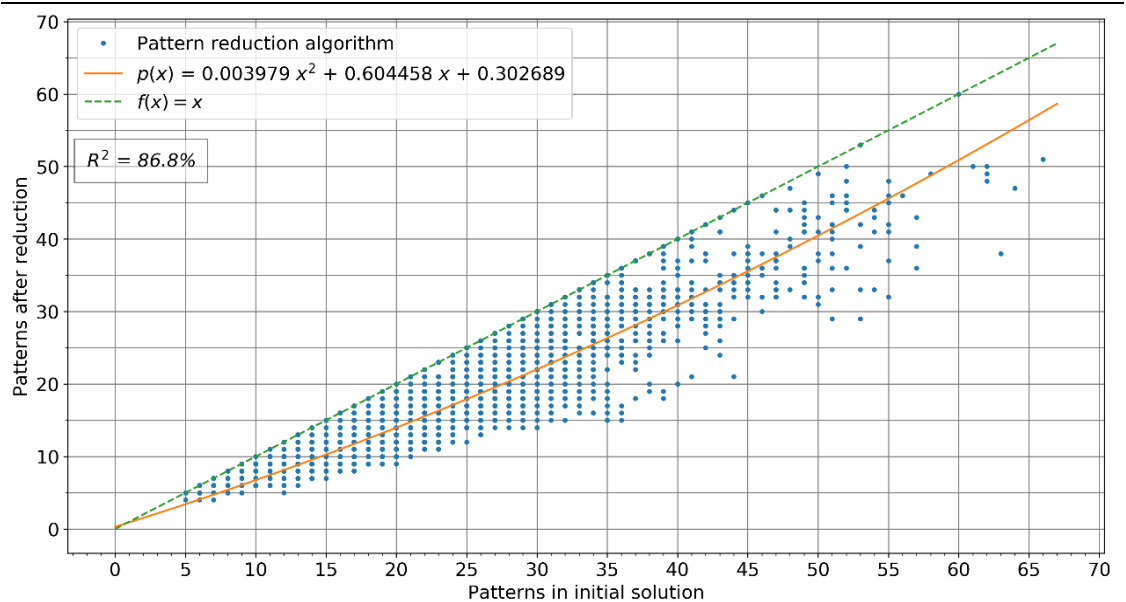

**Figure 2. Scatter diagram showing original vs. final pattern count on the 7,440 instances in the training data set.**

Using the quadratic on the remaining 1,860 (20%) instances we get a Mean Absolute Percentage Error (MAPE) of 12.0%.

## 4 The ML Model & Training

For the training, we used the same 80% of the data (7,440 instances, randomly chosen) that was used to fit the quadratic, keeping 20% for validation.

We used the popular open-source TensorFlow ML framework with the following configuration:

| Layer | Input (neurons) | Output (neurons) | Activation |
|---|---|---|---|
| **0** | (canonical solution) 10,802 | 100 | reLU |
| **1** | 100 | 100 | reLU |
| **2** | 100 | 1 | Linear |

**Optimiser**: Adam [10] with learning rate = 0.001

**Loss function**: Mean Absolute Error (MAE)

**Epochs**: 500

**Stopping criterion:** 25 epochs without improvement in the validation set (20%)

The model with the lowest MAE for the validation set is the one saved and used for testing. The training always ends before epoch 100. On the same computer as used for the data generation, the training takes about 2.5 minutes. Figure 3 shows a training session example.

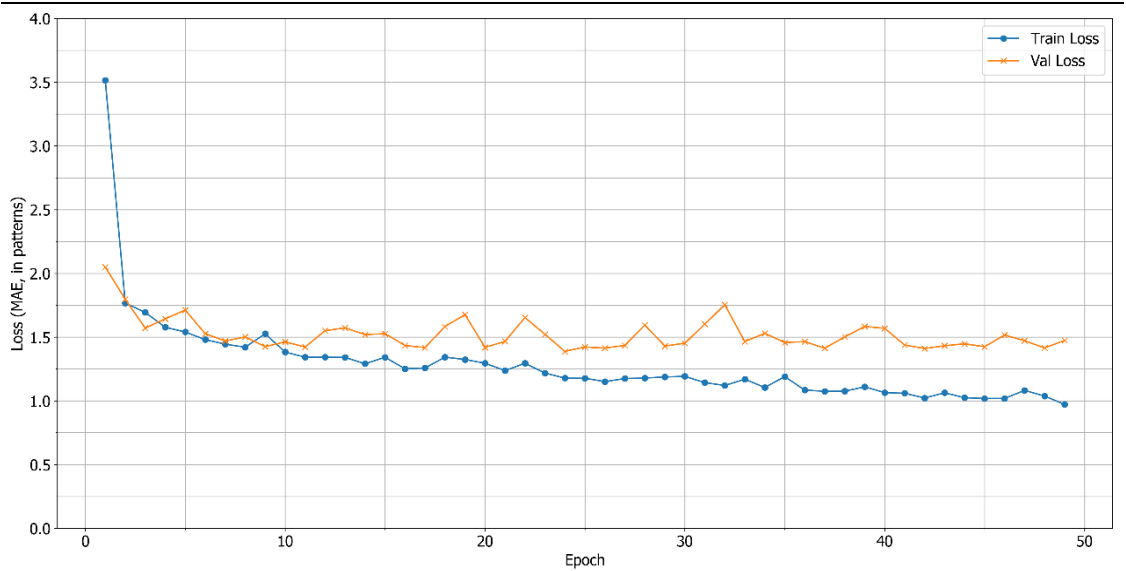

**Figure 3. Training example showing how the Mean Absolute Error evolves.**

As part of the process of selecting the ML model, we tried different optimisers. The table on the right shows the different results we obtained with each optimiser.

In addition, we tested different learning rates (0.0005 to 0.002), number of hidden layers (1 to 3), dropout layers and neurons per layer (32, 64, 128 and 256). The difference in the results from the chosen configuration was always a small increase of the MAPE.

| Optimiser | MAPE | Std. Dev. |
| --- | --- | --- |
| Adam | 8.74 | 0.138 |
| Adamax | 8.81 | 0.138 |
| Adagrad | 8.97 | 0.131 |
| Nadam | 9.05 | 0.118 |
| RMSprop | 9.13 | 0.555 |
| Adadelta | 9.26 | 0.171 |

## 5   Results & Discussion

The ML model on the testing data set generates, on average, a MAPE of 8.7%, which compares quite favourably with the average 12.0% of the naïve model.

The trained ML model produces an answer in a trivial time, < 1 ms per instance.

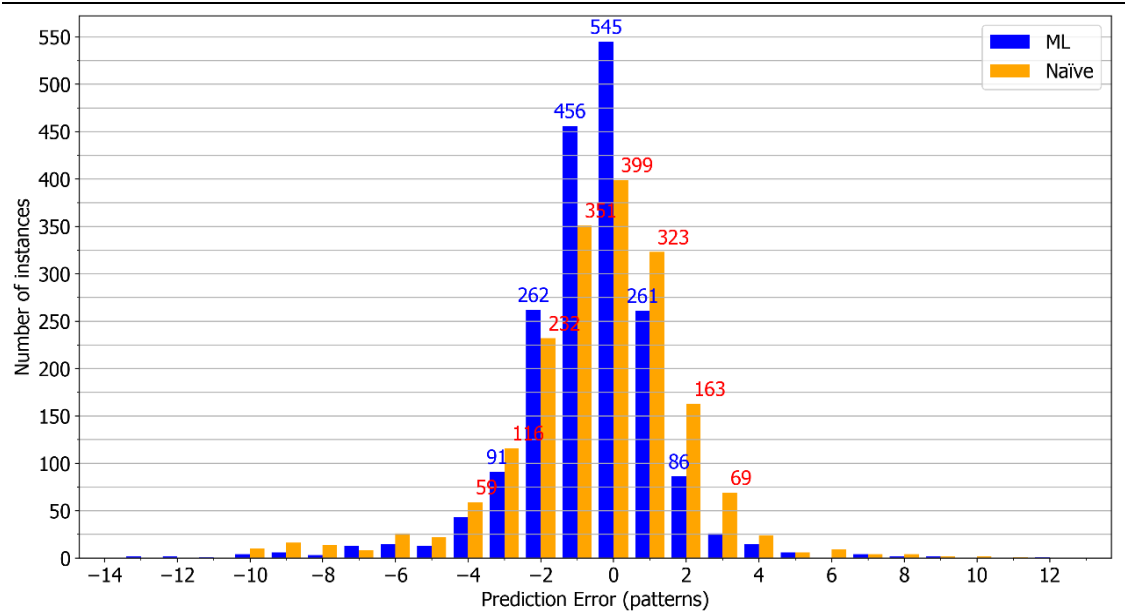

**Figure 4. Error histogram for the predictions obtained from the ML and Naïve models.**

A negative prediction error means the predicted value was smaller than the actual.

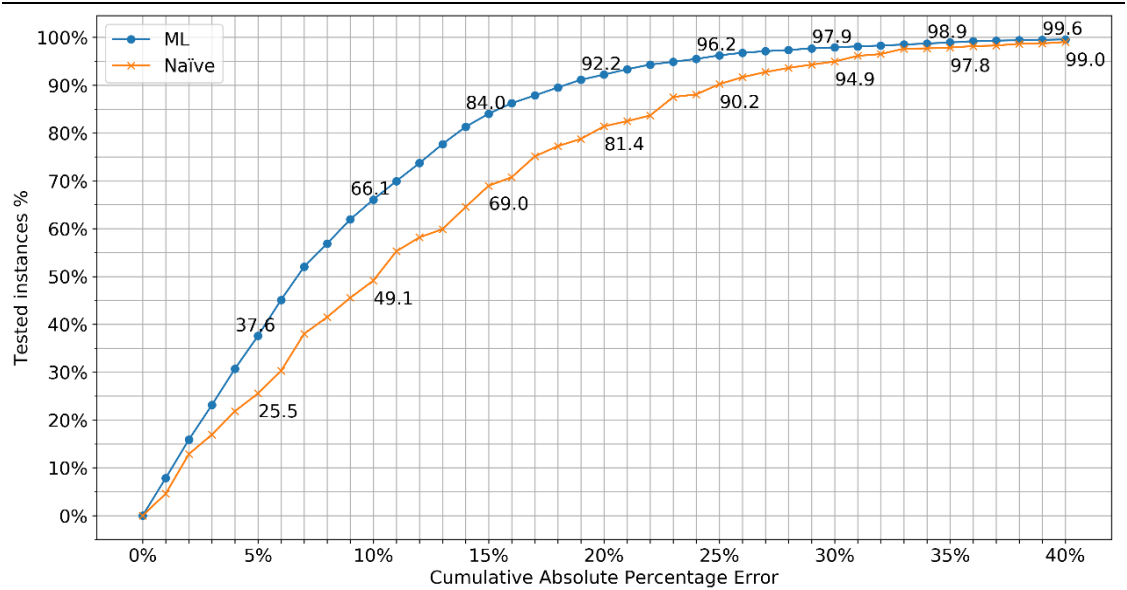

**Figure 5. Comparison of absolute errors between naïve and ML models.**

The coefficient of determination of the ML model (91.7%) shows that the prediction accuracy is better than the naïve approach (86.8%).

Incidentally, the ML model predicts 16.7 patterns when passed the solution to the left in Figure 1, whereas the naïve model predicts 19.5.

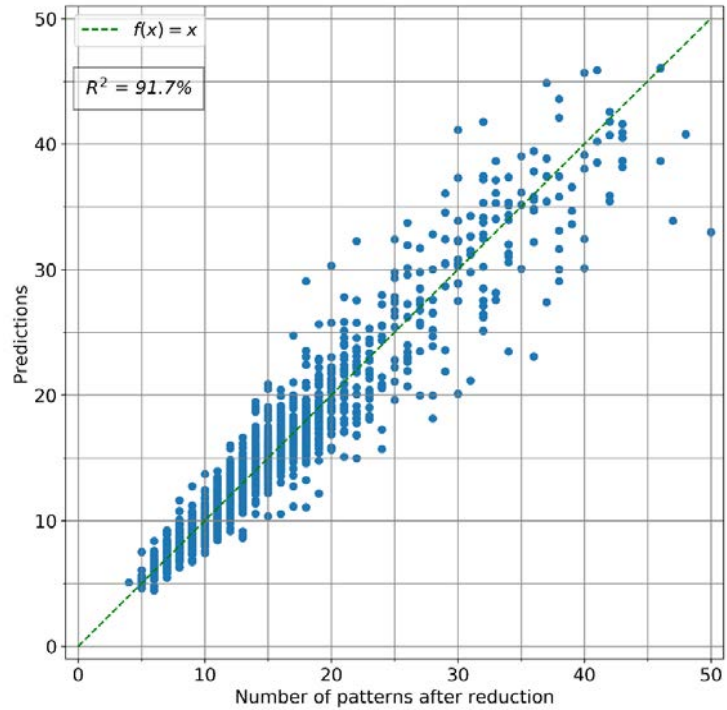

**Figure 6. Scatter diagram showing final pattern count vs. predicted by the ML model on the 1,860 instances of the testing set.**

Following this work, there are many open questions.

One of them concerns the integrality of the answer. Both the naïve and the ML algorithm return a floating point number, yet the number of patterns in a solution remains resolutely integer. There are no clear arguments on whether rounding to the nearest or up/down is better.

Another question is whether we gained any insights into how the ML model was coming up with its predictions. Unfortunately we have no such insight at this point.

But there was a side-benefit: as mentioned earlier, in ~3% of instances our current algorithm failed to find an improvement in 150 s. Closer examination of these instances is leading to algorithmic improvements, which will be the subject of a future paper.

In terms of future work, one possible direction would be to "move further up the chain", namely to see how accurately can ML predict solution statistics (waste, pattern count,…) from the input of the 1D-CSP problem itself (list of sizes, master size(s), constraints).

# 6 Acknowledgments

The authors would like to acknowledge constructive comments by Ed Rothberg and Sophia Drossopoulou on an earlier draft.

# 7 References

[1] C. Goulimis and A. Olivera, "KBP: A New Pattern Reduction Heuristic for the Cutting Stock Problem," in *OR59*, Loughborough, 2017.

[2] C. McDiarmid, "Pattern Minimisation in Cutting Stock Problems," *Discrete Applied Mathematics,* pp. 121-130, 1999.

[3] C. N. Goulimis, "Appeal to NP-Completeness Considered Harmful: Does the Fact That a Problem Is NP-Complete Tell Us Anything?," *Interfaces,* pp. 584-586, 2007.

[4] C. N. Goulimis, "Optimal solutions for the cutting stock problem," *European Journal of Operational Research,* pp. 197-208, 1990.

[5] F. Vanderbeck, "Exact Algorithm for Minimising the Number of Setups in the One-Dimensional Cutting Stock Problem," *Operations Research,* pp. 915-926, 2000.

[6] G. Belov and G. Scheithauer, "The number of setups (different patterns) in one-dimensional stock cutting," Department of Mathematics, Dresden University of Technology, Dresden, 2003.

[7] Y. Cui, C. Zhong and Y. Yao, "Pattern-set generation algorithm for the one-dimensional cutting stock problem with setup cost," *European Journal of Operational Research,* pp. 540-546, 2015.

[8] Y. Bengio, A. Lodi and A. Prouvost, "Machine Learning for Combinatorial Optimization: a Methodological Tour d'Horizon," *CoRR,* 2018.

[9] K. He, X. Zhang, S. Ren and J. Sun, "Spatial Pyramid Pooling in Deep Convolutional Networks for Visual Recognition," *CoRR,* 2014.

[10] D. Kingma and J. Ba, "Adam: A method for stochastic optimization," in *3rd International Conference for Learning Representations*, San Diego, 2015.